\documentclass[sn-basic,iicol]{sn-jnl}  

\usepackage{graphicx}%
\usepackage{multirow}%
\usepackage{amsmath,amssymb,amsfonts}%
\usepackage{amsthm}%
\usepackage{mathrsfs}%
\usepackage[title]{appendix}%
\usepackage{xcolor}%
\usepackage{textcomp}%
\usepackage{manyfoot}%
\usepackage{booktabs}%
\usepackage{algorithm}%
\usepackage{algorithmicx}%
\usepackage{algpseudocode}%
\usepackage{listings}%
\usepackage{subcaption}
\usepackage{multirow}
\usepackage{amsmath}
\usepackage{bbding}
\usepackage{amssymb}
\usepackage{times}
\usepackage{bm}
\usepackage{lmodern}
\usepackage{anyfontsize}

\begin{document}

\title[Article Title]{Multi-Stream Keypoint Attention Network for Sign Language Recognition and Translation}


\author[1]{\fnm{Mo} \sur{Guan}}

\author*[1]{\fnm{Yan} \sur{Wang}}\email{wangyan@smail.sut.edu.cn}

\author[2]{\fnm{Guangkun} \sur{Ma}}

\author[1]{\fnm{Jiarui} \sur{Liu}}

\author[1]{\fnm{Mingzu} \sur{Sun}}

\affil[1]{\orgdiv{School of Information Science and Engineering}, \orgname{Shenyang University of Technology}, \orgaddress{\street{ShenLiao West Road}, \city{Shenyang}, \postcode{110870}, \state{Liaoning}, \country{China}}}

\affil[2]{\orgdiv{School of Software}, \orgname{Shenyang University of Technology}, \orgaddress{\street{ShenLiao West Road}, \city{Shenyang}, \postcode{110870}, \state{Liaoning}, \country{China}}}



\abstract{Sign language serves as a non-vocal means of communication, transmitting information and significance through gestures, facial expressions, and bodily movements. The majority of current approaches for sign language recognition (SLR) and translation rely on RGB video inputs, which are vulnerable to fluctuations in the background. Employing a keypoint-based strategy not only mitigates the effects of background alterations but also substantially diminishes the computational demands of the model. Nevertheless, contemporary keypoint-based methodologies fail to fully harness the implicit knowledge embedded in keypoint sequences. To tackle this challenge, our inspiration is derived from the human cognition mechanism, which discerns sign language by analyzing the interplay between gesture configurations and supplementary elements. We propose a multi-stream keypoint attention network to depict a sequence of keypoints produced by a readily available keypoint estimator. In order to facilitate interaction across multiple streams, we investigate diverse methodologies such as keypoint fusion strategies, head fusion, and self-distillation. The resulting framework is denoted as MSKA-SLR, which is expanded into a sign language translation (SLT) model through the straightforward addition of an extra translation network. We carry out comprehensive experiments on well-known benchmarks like Phoenix-2014, Phoenix-2014T, and CSL-Daily to showcase the efficacy of our methodology. Notably, we have attained a novel state-of-the-art performance in the sign language translation task of Phoenix-2014T. The code and models can be accessed at: \url{https://github.com/sutwangyan/MSKA}.}




\maketitle

\section{Introduction}\label{sec:intro}

\begin{figure*}[!t]
\centering
\includegraphics[width=160mm]{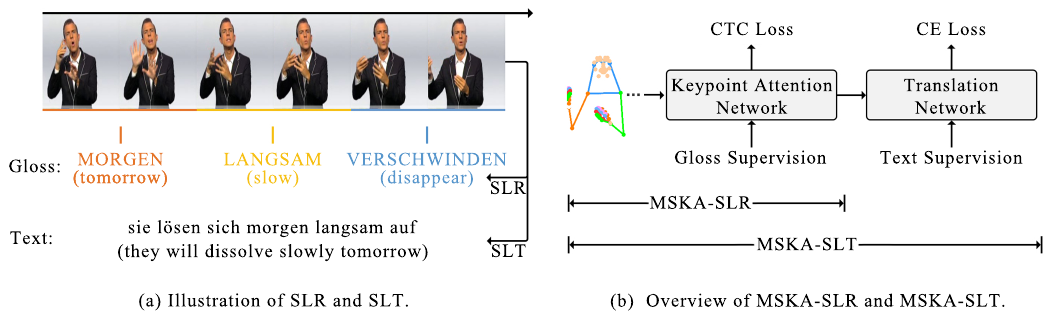}
\caption{(a)We choose a sign language video from the Phoenix-2014T dataset and display its gloss sequence alongside the corresponding text. The objective of sign language recognition (SLR) is to instruct models in producing matching gloss representations derived from sign language videos. Conversely, the task of sign language translation (SLT) entails creating textual representations that align with sign language videos. (b) MSKA-SLT is constructed on the foundation of MSKA-SLR to facilitate SLT. Keypoint sequences are depicted in coordinate form.
}
\label{fig1}
\end{figure*}

Sign language, a form of communication utilizing gestures, expressions, and bodily movements, has been the subject of extensive study~\cite{bungeroth2004statistical,starner1998real,tamura1988recognition}. For the deaf and mute community, sign language serves as their primary mode of communication. It holds profound significance, offering an effective medium for this particular demographic to convey thoughts, emotions, and needs, thereby facilitating their active participation in social interactions. Sign language possesses a unique structure, incorporating elements such as the shape, direction, and placement of gestures, along with facial expressions. Its grammar diverges from that of spoken language, exhibiting differences in grammatical structure and sequence. To address such disparities, certain sign language translation (SLT) tasks integrate gloss sequences before text generation. The transition from visual input to gloss sequences constitutes the process of sign language recognition (SLR). Fig.~\ref{fig1}(a) depicts both SLR and SLT tasks.

Gestures play a pivotal role in the recognition and translation of sign language. Indeed, gestures occupy a modest portion of the video, rendering them vulnerable to shifts in the background and swift hand movements during sign language communication. Consequently, this results in challenges in acquiring sign language attributes. Nevertheless, owing to robustness and computational efficiency of gestures, some methodologies advocate for the employment of keypoints to convey it. Ordinarily, sign language videos undergo keypoint extraction using off-the-shelf keypoint estimator. Following this, the keypoint sequences are regionally cropped to be utilized as input for the model, allowing a more precise focus on the characteristics of hand shapes. The TwoStream method, as described in~\cite{chen2022two}, enhances feature extraction by converting keypoints into heatmaps and implementing 3D convolution. SignBERT+, detailed in the work by~\cite{hu2023signbert+}, represents hand keypoints as a graphical framework and employs graph convolutional networks for extracting gesture features. Nevertheless, a key drawback of these approaches is the inadequate exploitation of correlation data among keypoints.

To address this challenge, we introduce an innovative network framework that depends entirely on the interplay among keypoints to achieve proficiency in sign language recognition and translation endeavors. Our methodology is influenced by the innate human inclination to prioritize the configuration of gestures and the dynamic interconnection between the hands and other bodily elements in the process of sign language interpretation. The devised multi-stream keypoint attention (MSKA) mechanism is adept at facilitating sign language translation by integrating a supplementary translation network. As a result, the all-encompassing system is designated as MSKA-SLT, as illustrated in Fig.~\ref{fig1}(b).

In summary, our contributions primarily consist of the following three aspects:
\begin{enumerate}
\item To the best of our knowledge, we are the first to propose a multi-stream keypoint attention, which is built with pure attention modules without manual designs of traversal rules or graph topologies.
\item We propose to decouple the keypoint sequences into four streams, left hand stream, right hand stream, face stream and whole body stream, each focuses on a specific aspect of the skeleton sequence. By fusing different types of features, the model can have a more comprehensive understanding for sign language recognition and translation.
\item We conducted extensive experiments to validate the proposed method, demonstrating encouraging improvements in sign language recognition tasks on the three prevalent benchmarks, \emph{i.e.}, Phoenix-2014~\cite{koller2015continuous}, Phoenix-2014T~\cite{camgoz2018neural} and CSL-Daily~\cite{zhou2021improving}. Moreover, we achieved new state-of-the-art performance in the translation task of Phoenix-2014T.
\end{enumerate}

\section{Related Work}

\subsection{Sign Language Recognition and Translation}

Sign language recognition is a prominent research domain in the realm of computer vision, with the goal of deriving sign glosses through the analysis of video or image data. 2D CNNs are frequently utilized architectures in computer vision to analyze image data, and they have garnered extensive use in research pertaining to sign language recognition~\cite{cihan2017subunets,niu2020stochastic,cui2019deep,zhou2021spatial,hu2023continuous,hu2023adabrowse,guo2023distilling}.

STMC~\cite{zhou2021spatial} proposed a spatio-temporal multi-cue network to address the problem of visual sequence learning. CorrNet~\cite{hu2023continuous} model captures crucial body movement trajectories by analyzing correlation maps between consecutive frames. It employs 2D CNNs to extract image features, followed by a set of 1D CNNs to acquire temporal characteristics. AdaBrowse~\cite{hu2023adabrowse} introduced a novel adaptive model that dynamically selects the most informative subsequence from the input video sequence by effectively utilizing redundancy modeled for sequential decision tasks. CTCA~\cite{guo2023distilling} build a dual-path network that contains two branches for perceptions of local temporal context and global temporal context. By extending 2D CNNs along the temporal dimension, 3D CNNs can directly process spatio-temporal information in video data. This approach enables a better understanding of the dynamic features of sign language movements, thus enhancing recognition accuracy~\cite{li2020tspnet,pu2019iterative,chen2022simple}. MMTLB~\cite{chen2022simple} utilize a pre-trained S3D~\cite{xie2018rethinking} network to extract features from sign language videos for sign language recognition, followed by the use of a translation network for sign language translation tasks. Recent studies in gloss decoder design have predominantly employed either Hidden Markov Models (HMM)~\cite{koller2017re,koller2018deep,koller2019weakly} or Connectionist Temporal Classification (CTC)~\cite{cheng2020fully,Min_2021_ICCV,zhou2021spatial}, drawing from their success in automatic speech recognition. We opted for CTC due to its straightforward implementation. While CTC loss offers only modest sentence-level guidance, approaches such as those proposed by ~\cite{cui2019deep,zhou2019dynamic,chen2022two} suggest iteratively deriving detailed pseudo labels from CTC outputs to enhance frame-level supervision. Additionally,~\cite{Min_2021_ICCV} achieves frame-level knowledge distillation by aligning the entire model with the visual encoder. 

In this study, our distillation process leverages the multi-stream architecture to incorporate ensemble knowledge into each individual stream, thereby improving interaction and coherence among the multiple streams. Sign language translation (SLT) involves directly generating textual outputs from sign language videos. Many existing methods frame this task as a neural machine translation (NMT) challenge, employing a visual encoder to extract visual features and feeding them into a translation network for text generation ~\cite{camgoz2018neural,camgoz2020sign,chen2022simple,li2020tspnet,zhou2021improving,xie2018rethinking,chen2022two}. We adopt mBART ~\cite{liu2020multilingual} as our translation network, given its impressive performance in SLT~\cite{chen2022simple,chen2022two}. To attain satisfactory outcomes, gloss supervision is commonly employed in SLT. This involves pre-training the multi-stream attention network on SLR ~\cite{camgoz2020sign, zhou2021spatial, zhou2021improving} and jointly training SLR and SLT ~\cite{zhou2021spatial, zhou2021improving}.

\begin{figure*}[!t]
\centering
\includegraphics[width=160mm]{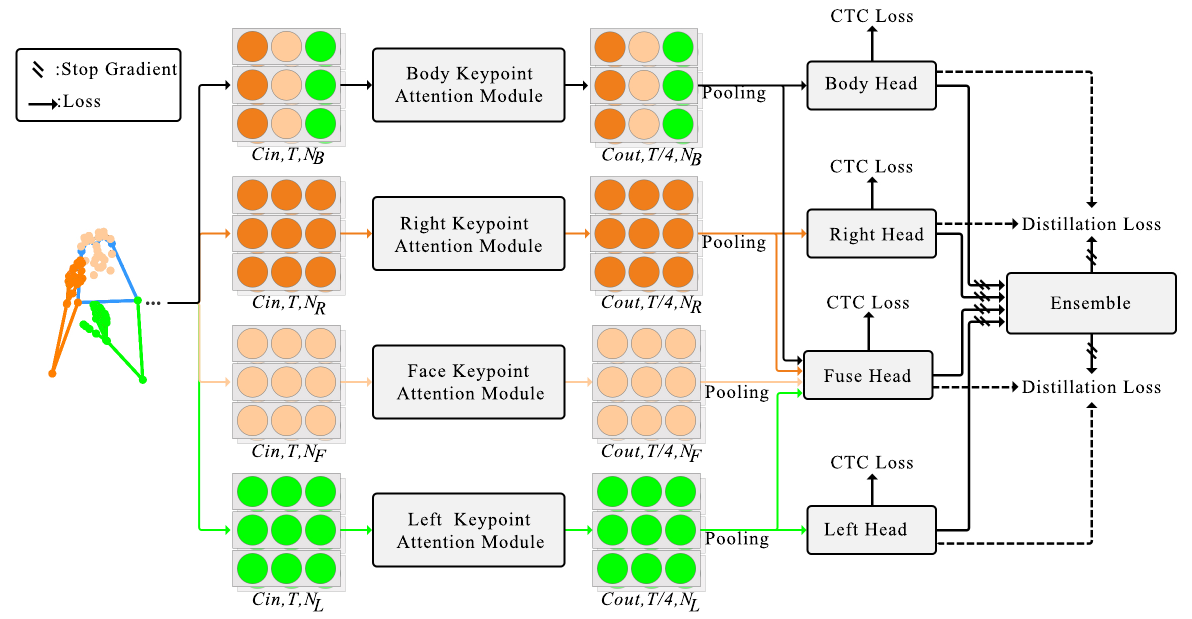}
\caption{The overview of our MSKA-SLR. The whole network is jointly supervised by the CTC losses and the self-distillation losses. Keypoints are represented in coordinate format.
}
\label{fig2}
\end{figure*}

\subsection{Introduce Keypoints into SLR and SLT}

The optimization of keypoints to enhance the efficacy of SLR and SLT remains a challenging issue. \cite{camgoz2020multi} introduce an innovative multichannel transformer design. The suggested structure enables the modeling of both inter and intra contextual connections among distinct sign articulators within the transformer network, while preserving channel-specific details. \cite{papadimitriou2020multimodal} presenting an end-to-end deep learning methodology that depends on the fusion of multiple spatio-temporal feature streams, as well as a fully convolutional encoder-decoder for prediction. TwoStream-SLR~\cite{chen2022two} put forward a dual-stream network framework that integrates domain knowledge such as hand shapes and body movements by modeling the original video and keypoint sequences separately. It utilizes existing keypoint estimators to generate keypoint sequences and explores diverse techniques to facilitate interaction between the two streams. SignBERT+~\cite{hu2023signbert+} incorporates graph convolutional networks (GCN) into hand pose representations and amalgamating them with a self-supervised pre-trained model for hand pose, the aim is to enhance sign language understanding performance. This method utilizes a multi-level masking modeling approach (including joint, frame, and clip levels) to train on extensive sign language data, capturing multi-level contextual information in sign language data. C$^2$SLR~\cite{zuo2024improving} aims to ensure coherence between the acquired attention masks and pose keypoint heatmaps to enable the visual module to concentrate on significant areas.

\begin{figure*}[!ht]
    \centering
    \includegraphics[width=160mm]{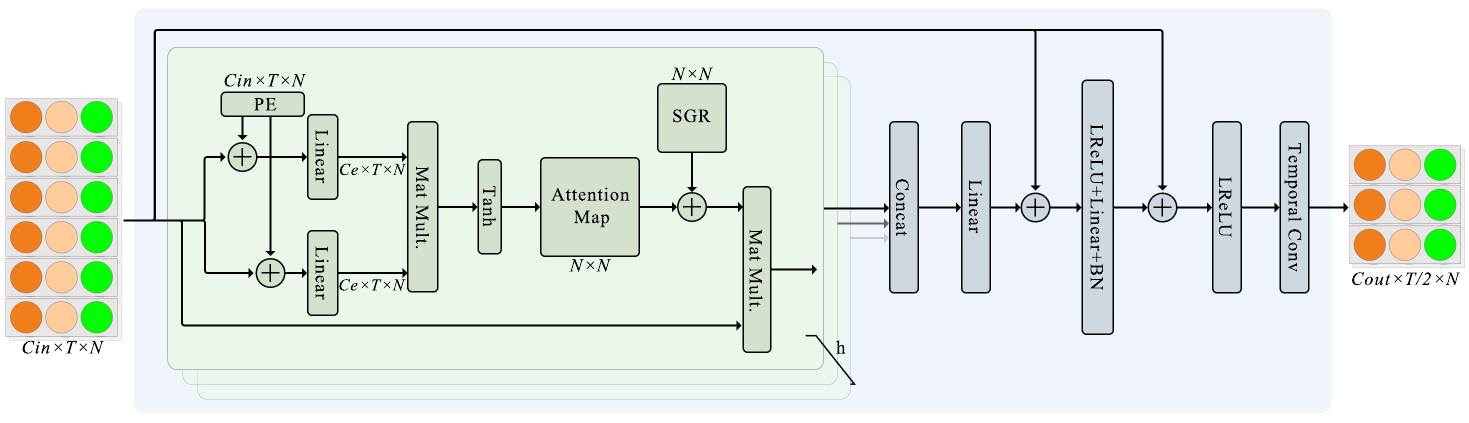}
    \caption{Illustration of the attention module. We show the body attention module as an example. The others attention module is an analogy. The green rounded rectangle box represents a single-head self-attention module. There are totally $h$ self-attention heads, whose output are concatenated and fed into two linear layers to obtain the output. LReLU represents the leaky ReLU~\cite{maas2013rectifier}.
    }
    \label{fig3}
\end{figure*}

\subsection{Self-attention mechanism}

Serves as the foundational component within the transformer architecture~\cite{vaswani2017attention,dai2019transformer}, representing a prevalent approach in the realm of natural language processing (NLP). Its operational framework encompasses a set of queries $Q$, keys $K$, and values $V$, each with a dimensionality of $C$, arranged in matrix format to facilitate efficient computation. Initially, the mechanism computes the dot product between the queries and all keys, subsequently normalizing each by $\sqrt{C}$ and applying a softmax function to derive the corresponding weights assigned to the values~\cite{vaswani2017attention}. Mathematically, this process can be formulated as follows:
\begin{equation}
\begin{aligned}
\label{equ:1}
\ Attention(Q,K,V) = softmax(\frac{ QK^{T}} {\sqrt{C}})V
\end{aligned}
\end{equation}

\subsection{Multi-Stream Networks}
In this work, our approach directly models keypoint sequences through an attention module. Additionally, to mitigate the issue of data scarcity and better capture glosses across different body parts, we introduce multi-stream attention to drive meaningful feature extraction of local features. Modeling the interactions among distinct streams presents a challenging challenge. I3D~\cite{carreira2017quo} adopts a late fusion strategy by simply averaging the predictions of the two streams. Another approach involves early fusion by lateral connections~\cite{feichtenhofer2019slowfast}, concatenation~\cite{zhou2021spatial}, or addition~\cite{cui2019deep} to merge intermediary features of each stream. In this study, we utilize the concept of lateral connections to facilitate mutual supplementation between multiple streams. Additionally, our self-distillation method integrates knowledge from multiple streams into the generated pseudo-targets, thereby achieving a more profound interaction.

\section{Proposed Method}

In this section, we initially present the data augmentation techniques for keypoint sequences. Subsequently, we elaborate on the individual components of MSKA-SLR. Finally, we outline the composition of MSKA-SLT.

    \subsection{Keypoint augment}

    Typically, sign language video datasets are constrained in size, underscoring the importance of data augmentation. In contrast to prior works such as~\cite{guo2023distilling,hu2023adabrowse,hu2023continuous,chen2022two}, our input data comprises keypoint sequences. Analogous to the augmentation techniques employed in image-related tasks, we implement a step for keypoints: Utilizing HRNet~\cite{wang2020deep} to extract keypoints from sign language videos, wherein the keypoint coordinates are denoted with respect to the top-left corner of the image, with the positive $X$ and $Y$ axes oriented towards the rightward and downward directions, respectively. To utilize data augmentation, we pull the origin back to the center of the image and normalize it by a function: $(( x / W , ( H - y ) / H ) - 0.5 ) / 0.5$ , with horizontal to the right and vertical upwards defining the positive directions of the $X$ and $Y$ axes, respectively. Within this context, the variables $x$ and $y$ denote the coordinates of a given point, whereas $H$ and $W$ symbolize the height and width of the image, respectively.
    
    1) We adjust the temporal length of the keypoint sequences within the interval [$\times0.5$-$\times 1.5$], selecting valid frames randomly from this range. 
    2) The scaling process involves multiplying the coordinates of each point in the provided keypoint set by a scaling factor.
    3) The transformation operation is implemented by applying the provided translation vector to the coordinates of each point in the provided set of keypoint coordinates.
    4) During the process of rotation, we achieve this by creating a matrix representing the rotation angle. Given a point $P(x,y)$ in two dimensions, the formula for calculating for the resulting point $P'(x',y')$ with the center at the origin, and a counterclockwise rotation by an angle of $\theta$, is as follows:
    \begin{equation}
    \begin{aligned}
    \label{equ:2}
    \begin{bmatrix} x ^ { \prime } \\ y ^ { \prime } \end{bmatrix} = \begin{bmatrix} \cos ( \theta ) ,  & - \sin ( \theta ) \\ \sin ( \theta ) , & \cos ( \theta ) \end{bmatrix} \begin{bmatrix} x \\ y \end{bmatrix}
    \end{aligned}
    \end{equation}
    Where $\cos ( \theta )$ and $\sin ( \theta )$ are respectively the cosine and sine values of the rotation angle $\theta$. 	The matrix multiplication operation rotates a two-dimensional point at coordinates $(x,y)$ counterclockwise around the origin point by an angle of $\theta$, yielding the rotated point $(x',y')$.

    \subsection{SLR}

    \subsubsection{Keypoint decoupling}

    We noted that the various components of the keypoint sequences within the same sign language sequence should convey the same semantic information. Thus, we divide the keypoint sequences into four sub-sequences: left hand, right hand, facial expressions and overall, and process them independently. Markers of different colors represent distinct keypoint sequences, as illustrated in Fig~\ref{fig2}. This segmentation helps the model more accurately capture the relationships between different parts, facilitating the provision of richer diversity of information. By handling them separately, the model can more attentively capture their respective key features. This keypoint decoupling strategy result and enhances over SLR predictions as shown in our experiments. 
    
    \subsubsection{Keypoint attention module}

    We employ HRNet~\cite{wang2020deep}, which has been trained on COCO-WholeBody~\cite{jin2020whole} dataset, to generate $133$ keypoints, including hand, mouth, eye, and body trunk keypoints. Consistent with \cite{chen2022two}, we employ a subset of $79$ keypoints, comprising $42$ hand keypoints, $11$ upper body keypoints covering shoulders, elbows, and wrists, and a subset of facial keypoints ($10$ mouth keypoints and $16$ others). Concretely, denoting the keypoint sequence as a multidimensional array with dimensions $C\times T\times N$, where the elements of $C$ consist of $[x_t^n,y_t^n,c_t^n]$, $(x_t^n,y_t^n)$ and $c_t^n$ denotes the coordinates and confidence of the $n$-th keypoint in the $t$-th frame, $T$ denotes the frame number, and $N$ is the total number of keypoints.
    
    As the attention modules for each stream are analogous, we choose the body keypoint attention module as an example for detailed elucidation. The complete attention module is depicted in the Fig.~\ref{fig3}. The procedure within the green rounded rectangle outlines the process of single-head attention computation. The input $ \bm{X}\in\mathbb{R}^{C\times T\times N}$ is first enriched with spatial positional encodings. It is then embedded into two linear mapping functions to obtain $ \bm{X}\in\mathbb{R}^{C_e\times T\times N}$, where ${C_e}$ is usually smaller than ${C_{out}}$ to alleviate feature redundancy and reduce computational complexity. The attention map is subjected to spatial global normalization. Note that when computing the attention map, we use the Tanh activation function instead of the softmax used in ~\cite{vaswani2017attention}. This is because the output of Tanh is not restricted to positive values, thus allowing for negative correlations and providing more flexibility~\cite{shi2020decoupled}. Finally, the attention map is element-wise multiplied with the original input to obtain the output features. 

    To facilitate the model to jointly attend to information from different representation subspaces, the module performs attention computation with $h$ heads. The outputs of all heads are concatenated and mapped to the output space. Similar to the~\cite{vaswani2017attention}, we add a feedforward layer at the end to generate the final output. We choose to use leaky ReLU~\cite{maas2013rectifier} as the non-linear activation function. Additionally, the module includes two residual connections to stabilize network training and integrate different features, as illustrated in the Fig.~\ref{fig3}. Finally, we employ 2D convolution to extract temporal features. All processes within the blue rounded rectangle constitute a complete keypoint attention module. It is worth noting that the weights of different keypoint attention modules are not shared.
    
    \subsubsection{Position encoding} 

    	The keypoint sequences are structured into a tensor and inputted to the neural network. Because there is no predetermined sequence or structure for each element of the tensor, we require a positional encoding mechanism to provide a unique label for every joint. Following~\cite{vaswani2017attention,shi2020decoupled}, we employ sinusoidal and cosine functions with different frequencies as encoding functions:
	    \begin{equation}
	    \begin{aligned}
	    \label{equ:2}
	    \ PE ( p , 2 i ) = \sin ( p / 10000^{2i/C_{in } } ) \\ PE ( p , 2 i + 1 ) = \cos ( p / 10000^{2i/C_{in} } )
	    \end{aligned}
	    \end{equation}
	    Where $p$ represents the position of the element and $i$ denotes the dimension of the positional encoding vector. Incorporating positional encoding allows the model to capture positional information of elements in the sequence. Their periodic nature provides different representations for distinct positions, enabling the model to better understand the relative positional relationships between elements in the sequence. Joints within a single frame are sequentially encoded, while identical joint across various frames shares a common encoding. It's worth noting that in contrast to the approach proposed in~\cite{shi2020decoupled}, we only introduce positional encoding for the spatial dimension. We use 2D convolution to extract temporal features, eliminating the need for additional temporal encoding as the continuity of time is already considered in the convolution operation.

    \subsubsection{Spatial global regularization}

    	For action detection tasks on skeleton data, the fundamental concept is to utilize known information, namely that each joint of the human body have unique physical or semantic attributes that remain invariant and consistent across all time frames and instances of data. Utilizing this known information, the objective of spatial global regularization is to encourage the model to grasp broader attention patterns, thus better adapting to diverse data samples. This method is achieved by implementing a global attention matrix, presented in the form of $N \times N$, showing the universal relationships among the body joints. This global attention matrix is shared across all data instances and optimized alongside other parameters during training of the network.

    \subsubsection{Head Network}
    The output feature from the final attention block undergoes spatial pooling to reduce its dimensions to ${T/4} \times 256$ before being inputted into the head network in the Fig.~\ref{fig2}. The primary objective of the head network is to further capture temporal context. It is comprised of a temporal linear layer, a batch normalization layer, a ReLU layer, along with a temporal convolutional block containing two temporal convolutional layers with a kernel size of $3$ and a stride of $1$, followed by a linear translation layer and another ReLU layer. The resulting feature, known as gloss representation, has dimensions of ${T/4} \times 512$. Subsequently, a linear classifier and a softmax function are utilized to extract gloss probabilities. We use connectionist temporal classification (CTC) loss $\mathcal{L}^{body}_{CTC}$ to optimize the body attention module.

    \subsubsection{Fuse Head and Ensemble}

    	Every keypoint attention module possesses a distinct array of network heads. To thoroughly harness the capabilities of our multi-stream architecture, we integrate an auxiliary fuse head, designed to assimilate outputs from various streams. This fusion head's configuration mirrors that of its counterparts, like the body head, and is likewise governed by CTC loss, represented as $\mathcal{L}^{fuse}_{CTC}$. The forecasted frame gloss probabilities are averaged and subsequently furnished to an ensemble to fabricate the gloss sequence. This ensemble approach amalgamates outcomes from multiple streams, thereby enhancing predictions, as demonstrated in the experiments.

    \subsubsection{Self-Distillation}

    	Frame-Level Self-Distillation~\cite{chen2022two} is employed, where the predicted frame gloss probabilities are used as pseudo-targets. In addition to coarse-grained CTC loss, extra fine-grained supervision is provided. Pursuant to our multi-stream design, we use the average gloss probability from the four head networks as pseudo-targets to guide the learning process of each stream. In a formal capacity, we endeavor to diminish the KL divergence between the pseudo-targets and the predictions of the four head networks. This process is designated as frame-level self-distillation loss, for it provides not merely frame-specific oversight but also filters insights from the concluding ensemble into each distinct stream.

    \subsubsection{Loss Function}\label{sec:slr_loss}

    The overall loss of MSKA-SLR is composed of two parts:1) the CTC losses applied on the outputs of the left stream($\mathcal{L}^{left}_{CTC}$), right stream($\mathcal{L}^{right}_{CTC}$), body stream($\mathcal{L}^{body}_{CTC}$), fuse stream($\mathcal{L}^{fuse}_{CTC}$); 2) the distillation loss ($\mathcal{L}_{Dist}$). We formulate the recognition loss as follows:
    \begin{equation}
    \begin{aligned}
    \label{equ:2}
    L _ { S L R } = L _ { C T C } ^ { l e f t } + L _ { C T C } ^ { r i g h t } + L _ { C T C } ^ { b o d y } + L _ { C T C } ^ { f u s e } + L _ { D i s t }
    \end{aligned}
    \end{equation}
    Up to now, we have introduced all components of MSKA-SLR. Once the training is finished, MSKA-SLR is capable of predicting a gloss sequence by fuse head network.

    \begin{table*}[!ht]
    \caption{Comparison with previous works on Sign Language Recognition (SLR). WER is adopted as the evaluation metric. Pre: pre-trained.}
    \begin{center}
    \resizebox{0.8\linewidth}{!}{
    \begin{tabular}{l c cc cc cc}
    \toprule
    \multirow{2}{*}{Method} & \multirow{2}{*}{Pre}  & \multicolumn{2}{c}{Phoenix-2014}
                            & \multicolumn{2}{c}{Phoenix-2014T}
                            & \multicolumn{2}{c}{CSL-Daily} \\

            & & Dev & Test
            & Dev & Test 
            & Dev & Test    \\
    \midrule
    \textbf{RGB-based} & & &  &  & & &  \\
    SubUNets~\cite{cihan2017subunets}       &\Checkmark & 40.8  & 40.7 & - & - & 41.4 & 41.0 \\
    LS-HAN~\cite{huang2018video}             &\Checkmark &  -    &  -   & - & - & 39.0 & 39.4 \\
    Hybrid CNN-HMM~\cite{koller2018deep}     &\Checkmark & 31.6  & 32.5 &-&-&-&-    \\ 
    DNF~\cite{cui2019deep}                   &\Checkmark & 23.8  & 24.4 & - & - & 32.8 & 32.4\\
    CNN-LSTM-HMM~\cite{koller2019weakly}     &\XSolidBrush & 26.0  & 26.0 & 22.1 & 24.1 & - & - \\
    FCN~\cite{cheng2020fully}                &\XSolidBrush & 23.7  & 23.9 & 23.3 & 25.1 &33.2&33.5\\
    Joint-SLRT~\cite{camgoz2020sign}		 &\XSolidBrush & - & - & 24.6 & 24.5 & 33.1 & 32.0 \\
    PiSLTRc-R~\cite{xie2021pisltrc}          &\Checkmark & 23.4  & 23.2 &-&-&-&-   \\
    SignBT~\cite{zhou2021improving}          &\Checkmark & -  & -  & 22.7 & 23.9 &33.2&33.2 \\  
    VAC~\cite{Min_2021_ICCV}                 &\Checkmark & 21.2  & 22.3 & -    & -    & - & - \\
    STMC~\cite{zhou2021spatial}              &\Checkmark & 21.7  & 20.7 & 19.6 & 21.0 & - & -  \\
    MMTLB~\cite{chen2022simple}              &\Checkmark &-&-      & 21.9 & 22.5 & -  & - \\ 
    C$^2$SLR~\cite{zuo2022c2slr}             &\Checkmark & 20.5  & 20.4 & 20.2 & 20.4 & - & -   \\
    CorrNet~\cite{hu2023continuous}          &\Checkmark & 18.8  & 19.4 & 18.9 & 20.5 & 30.6 & 30.1 \\
    TwoStream-SLR~\cite{chen2022two}         &\Checkmark & \textbf{18.4} & \textbf{18.8} &\textbf{17.7}&\textbf{19.3}&\textbf{25.4}&\textbf{25.3} \\
    SignBERT+~(+ R)~\cite{hu2023signbert+}   &\Checkmark & 19.9  & 20.0 & 18.8 & 19.9 & - & -  \\
    CTCA~\cite{guo2023distilling}            &\Checkmark & 19.5  & 20.1 & 19.3 & 20.3 & 31.3 & 29.4 \\
    AdaBrowse+~\cite{hu2023adabrowse}        &\Checkmark & 19.6  & 20.7 & 19.5 & 20.6 & 31.2 & 30.7 \\ 
    \midrule
    \textbf{Keypoint-based} & & & & & & &  \\
    TwoStream-SLR~\cite{chen2022two}& \Checkmark  & 28.6 & 28.0 & 27.1 & 27.2 & 34.6 & 34.1 \\
    SignBERT+~\cite{hu2023signbert+}& \Checkmark  & 34.0  &  34.1 & 32.9 & 33.6 & - & -  \\ 
    Ours~                           & \XSolidBrush & \textbf{21.7}  & \textbf{22.1} & \textbf{20.1} & \textbf{20.5} & \textbf{28.2} & \textbf{27.8} \\
    \bottomrule
    \end{tabular} }
    \end{center}
    \label{tab:slr}
\end{table*}

    \subsection{SLT}
    
    The traditional methodologies from previous times frequently described sign language translation (SLT) tasks as challenges in neural machine translation (NMT), where the input to the translation network is visual information. This research followed to this approach and implemented a multi-layer perceptron (MLP) with two hidden layers into the MSKA-SLR framework proposed, followed by the translation process, thereby accomplishing SLT. The network constructed in this manner is named MSKA-SLT, with its architecture illustrated in Fig.~\ref{fig1}(b). We chose to utilize employ mBART~\cite{liu2020multilingual} as the translation network due to its outstanding performance in cross-lingual translation tasks. To fully exploit the multi-stream architecture we designed, we appended an MLP and a translation network to the fuse head. The input to the MLP consists of encoded features by the fuse head network, namely the gloss representations. The translation loss is a standard sequence-to-sequence cross-entropy loss~\cite{vaswani2017attention}. MSKA-SLT includes the recognition loss Eq.~\ref{equ:2} and the translation loss represented by $L_{T}$, as specified in the formula:
    \begin{equation}
    \begin{aligned}
    \label{equ:3}
    L_{SLT} = L_{SLR} + L_{T} 
    \end{aligned}
    \end{equation}

\section{Experiments}

\paragraph{Implementation Details} To demonstrate the generalization of our methods, unless otherwise specified, we maintain the same configuration for all experiments. The network employs four streams, with each stream consisting $8$ attention blocks, and each block containing $6$ attention heads. The output channels are set as follows: $64, 64, 128, 128, 256, 256, 256$ and $256$ respectively. For SLR tasks, we utilize a cosine annealing schedule over $100$ epochs and an Adam optimizer with weight decay set to $1e-3$, and an initial learning rate of $1e-3$. The batch size is set to $8$. Following~\cite{chen2022simple,chen2022two}, we initialize our translation network with mBART-large-cc25\footnote[1]{https://huggingface.co/facebook/mbart-large-cc25} pretrained on CC25\footnote[2]{https://commoncrawl.org/}. We use a beam width of $5$ for both the CTC decoder and the SLT decoder during inference. We train for $40$ epochs with an initial learning rate of $1e-3$ for the MLP and $1e-5$ for MSKA-SLR and the translation network in MSKA-SLT. Other hyper-parameters remain consistent with MSKA-SLR. We train our models on one Nvidia 3090 GPU. 
    \begin{table*}[!t]
    \centering
    \caption{Performance comparison of MSKA-SLT with methods for SLT on Phoenix-2014T and CSL-Daily. }
    \resizebox{0.95\linewidth}{!}{
    \begin{tabular}{l c c c c c c c c c c}
    \toprule
    & \multicolumn{10}{c}{Phoenix-2014T}  \\
    \multirow{2}{*}{Methods} & \multicolumn{5}{c}{Dev} & \multicolumn{5}{c}{Test} \\ 
    & ROUGE & BLEU-1 & BLEU-2 & BLEU-3 & BLEU-4 & ROUGE & BLEU-1 & BLEU-2 & BLEU-3 & BLEU-4  \\ 
    \midrule
    \textbf{RGB-based} &  &  &  &  &  &  &  &  &  & \\
    Sign2Text~\cite{cihan2018neural} & 31.80 & 31.87 & 19.11 & 13.16 & 9.94 & 31.80 & 32.24 & 19.03 & 12.83 & 9.58 \\
    TSPNet~\cite{li2020tspnet} & - & - & - & - & - & 34.96 & 36.10 & 23.12 & 16.88 & 13.41 \\
    MCT~\cite{camgoz2020multi} & 45.90 & - & - & - & 19.51 & 43.57 & - & - & - & 18.51 \\
    SL-Trans~\cite{camgoz2020sign} & - & 47.26 & 34.40 & 27.05 & 22.38 & - & 46.61 & 33.73 & 26.19 & 21.32 \\
    BN-TIN-Trans~\cite{zhou2021improving} & 46.87 & 46.90 & 33.98 & 26.49 & 21.78 & 46.98 & 47.57 & 34.64 & 26.78 & 21.68 \\
    Joint-SLRT~\cite{camgoz2020sign} & - & 47.73 & 34.82 & 27.11 & 22.11 & - & 48.47 & 35.35 & 27.57 & 22.45 \\ 
    SimulSLT~\cite{yin2021simulslt} & 36.04 & 36.01 & 22.60 & 16.05 & 12.39 & 35.13 & 35.92 & 22.70 & 16.03 & 12.27 \\
    PiSLTRc-T~\cite{xie2021pisltrc} & 47.89 & 46.51 & 33.78 & 26.78 & 21.48 & 48.13 & 46.22 & 33.56 & 26.04 & 21.29 \\
    STMC~\cite{zhou2021spatial} & 48.24 & 47.60 & 36.43 & 29.18 & 24.08 & 46.65 & 46.98 & 36.09 & 28.70 & 23.65 \\
    SignBT~\cite{zhou2021improving} & 50.29 & 51.11 & 37.90 & 29.80 & 24.45 & 49.54 & 50.80 & 37.75 & 29.72 & 24.32 \\
    MMTLB~\cite{chen2022simple} & 53.10 & 53.95 & 41.12 & 33.14 & 27.61 & 52.65 & 53.97 & 41.75 & 33.84 & 28.39 \\
    TwoStream-SLT~\cite{chen2022two} & 54.08 & \textbf{54.32} & \textbf{41.99} & \textbf{34.15} & \textbf{28.66} & 53.48 & \textbf{54.90} & \textbf{42.43} & \textbf{34.46} & \textbf{28.95} \\ 
    ConSLT~\cite{fu2023token} & 47.52 & - & - & - & 24.27 & 47.65 & 51.57 & 38.81 & 30.91 & 25.48 \\ 
    SignBERT+(+ R)~\cite{hu2023signbert+} & 51.12 & 51.46 & 38.28 & 30.30 & 24.95 & 50.63 & 52.01 & 39.19 & 31.06 & 25.70 \\
    XmDA~\cite{ye2023cross} & 52.42 & - & - & - & 25.86 & 49.87 & - & - & - & 25.36 \\
    IP-SLT~\cite{yao2023sign} & \textbf{54.43} & 54.10 & 41.56 & 33.66 & 28.22 & \textbf{53.72} & 54.25 & 41.51 & 33.45 & 27.97 \\
    \midrule
    \textbf{Keypoint-based} &  &  &  &  &  &  &  &  &  & \\
    Skeletor~\cite{jiang2021skeletor} & 32.66 & 31.97 & 19.53 & 14.01 & 10.91 & 31.80 & 31.86 & 19.11 & 13.49 & 10.35 \\
    TwoStream-SLT~\cite{chen2022two} & \textbf{53.32} & 53.66 & \textbf{41.31} & \textbf{33.55} & \textbf{28.10} & 53.19 & 54.22 & 41.72 & 33.82 & 28.42 \\
    SignBERT+~\cite{hu2023signbert+} & 45.53 & 44.45 & 31.88 & 24.59 & 19.86 & 44.89 & 44.35 & 32.09 & 24.92 & 20.41 \\
    Ours~ & 52.67 & \textbf{54.09} & 41.29 & 33.24 & 27.63 & \textbf{53.54} & \textbf{54.79} & \textbf{42.42} & \textbf{34.49} & \textbf{29.03} \\ 
    \bottomrule
    \toprule
    & \multicolumn{10}{c}{CSL-Daily}  \\
    \multirow{2}{*}{Methods} & \multicolumn{5}{c}{Dev} & \multicolumn{5}{c}{Test} \\ 
    & ROUGE & BLEU-1 & BLEU-2 & BLEU-3 & BLEU-4 & ROUGE & BLEU-1 & BLEU-2 & BLEU-3 & BLEU-4  \\ 
    \midrule
    \textbf{RGB-based} &  &  &  &  &  &  &  &  &  & \\
    SL-Trans~\cite{camgoz2020sign} & 37.06 &  37.47 & 24.67 & 16.86 & 11.88 & 36.74 & 37.38 & 24.36 & 16.55 & 11.79 \\
    Joint-SLRT~\cite{camgoz2020sign} & 44.18 & 46.82 & 32.22 & 22.49 & 15.94 & 44.81 & 47.09 & 32.49 & 22.61 & 16.24 \\
    SignBT~\cite{zhou2021improving} & 49.49 & 51.46 & 37.23 & 27.51 & 20.80 & 49.31 & 51.42 & 37.26 & 27.76 & 21.34 \\
    MMTLB~\cite{chen2022simple} & 53.38 & 53.81 & 40.84 & 31.29 & 24.42 & 53.25 & 53.31 & 40.41 & 30.87 & 23.92 \\
    TwoStream-SLT~\cite{chen2022two} & \textbf{55.10} & \textbf{55.21} & \textbf{42.31} & \textbf{32.71} & \textbf{25.76} & \textbf{55.72} & \textbf{55.44} & \textbf{42.59} & \textbf{32.87} & \textbf{25.79} \\
    ConSLT~\cite{fu2023token} & 41.46 & - & - & - & 14.8 & 40.98 & - & - & - & 14.53 \\  
    XmDA~\cite{ye2023cross} & 49.36 & - & - & - & 21.69 & 49.34 & - & - & - & 21.58 \\
    IP-SLT~\cite{yao2023sign} & 44.33 & 45.26 & 31.77 & 22.87 & 16.74 & 44.09 & 44.85 & 31.50 & 22.66 & 16.72 \\
    \midrule
    \textbf{Keypoint-based} &  &  &  &  &  &  &  &  &  & \\
    TwoStream-SLT~\cite{chen2022two} & \textbf{54.03} & 54.43 & 41.60 & 31.95 & 25.01 & \textbf{55.07} & 55.34 & 42.36 & 32.58 & 25.42 \\
    Ours~ & 53.53 & \textbf{55.95} & \textbf{42.38} & \textbf{32.37} & \textbf{25.16} & 54.04 & \textbf{56.37} & \textbf{42.80} & \textbf{32.78} & \textbf{25.52} \\ 
	\bottomrule
  \end{tabular} }
  \vspace{1mm}
 \label{tab:slt}
 \vspace{-1mm}
 \end{table*}
    \subsection{Datasets and Evaluation Metrics}

    \subsubsection{Phoenix-2014}

    Phoenix-2014~\cite{koller2015continuous} is from weather forecast broadcasts aired on the German public TV station PHOENIX over a span of three years. This is a German SLR dataset with a vocabulary size of 1081 for glosses. The dataset comprises 5672, 540, and 629 instances in the training, development and testing set.
    
    \subsubsection{Phoenix-2014T}

    Phoenix-2014T~\cite{camgoz2018neural}is an extension of Phoenix-2014, has ascended as the foremost benchmark for SLR and SLT research in recent years~\cite{camgoz2018neural,simonyan2014two,mseqgraph,tran2015learning}. It encompasses an array of RGB sign language videos performed by a cadre of nine adept signers using German Sign Language (DGS). These videos are meticulously annotated with sentence-level glosses and accompanied by precise German translations transcribed from spoken news content. The dataset is methodically divided into training, development, and testing subsets, the dataset comprises $7096$, $519$, and $642$ video segments, respectively. With a vocabulary size of $1066$ for sign glosses and $2887$ for German text, Phoenix-2014T provides a rich resource for SLT research. With all ablation studies conducted using this comprehensive dataset.
    
    \subsubsection{CSL-Daily}

    CSL-Daily~\cite{zhou2021improving} is a recently released dataset for the translation of Chinese Sign Language (CSL), recorded in a studio environment. It encompasses $20654$ triplets of (video, gloss, text) enacted by ten unique signers. The dataset delves into diverse subjects such as familial existence, healthcare, and academic milieu. CSL-Daily is composed of $18401$, $1077$, and $1176$ partitions in the training, development and testing sections, correspondingly. The vocabulary size is $2000$ for sign glosses and $2343$ for Chinese text.

    \subsubsection{Evaluation Metrics}

    Following previous works~\cite{chen2022simple,zhou2021spatial,camgoz2020sign,camgoz2018neural,chen2022two,hu2023signbert+}, we adopt word error rate (WER) for SLR evaluation, and BLEU~\cite{papineni2002bleu} and ROUGE-L~\cite{lin2004rouge} to evaluate SLT. Lower WER indicates better recognition performance. For BLEU and ROUGE-L, the higher, the better. 

    \subsection{Comparison with State-of-the-art Methods}

    In this section, we compare our method with previous state-of-the-art methods on two main downstream tasks, including SLR and SLT. For comparison, we group them into RGB-based and Keypoint-based methods. 
    
    For SLR, we compare our recognition network with state-of-the-art methods on Phoenix-2014, Phoenix-2014T and CSL-Daily, as shown in Table~\ref{tab:slr}. The MSKA-SLR achieves $22.1$\%, $20.5$\% and $27.8$\% WER on the test sets of these three datasets, respectively. Typically, keypoint-based approaches are significantly falling short of RGB-based methods; however, our MSKA-SLR has substantially reduced this disparity. Among keypoint-based methods, our method significantly surpasses the most challenging competitor TwoStream-SLR~\cite{chen2022two} with $5.9$\%, $6.7$\% and $6.3$\% WER improvement on the testing sets of these three datasets, respectively. Note TwoStream-SLR~\cite{chen2022two} and SignBERT+~\cite{hu2023signbert+} utilize pre-trained model that leverage more model parameters and additional resources than MSKA-SLR. 
    \begin{table*}[!t]
    \centering
    \caption{Study the effects of each component of MSKA-SLR on the Phoenix-2014T SLR task.}
    \begin{tabular}{cccccccc}
    \toprule 
    Body & Left & Face & Right & Fuse Head & Distillation & Dev & Test \\
    \midrule
    \checkmark  &            &            &            &  &  & 25.30 & 25.50 \\
                & \checkmark &            &            &  &  & 57.37 & 57.60 \\
                & \checkmark & \checkmark &            &  &  & 34.90 & 36.26 \\
                &            &            & \checkmark &  &  & 35.28 & 35.58 \\
                &            & \checkmark & \checkmark &  &  & 25.86 & 26.05 \\
                & \checkmark & \checkmark & \checkmark &  &  & 25.27 & 25.69 \\ 
                & \checkmark & \checkmark & \checkmark & \checkmark &  & 23.08 & 23.67 \\
    \checkmark  & \checkmark & \checkmark & \checkmark & \checkmark &  & 22.69 & 22.70 \\
    \checkmark  & \checkmark & \checkmark & \checkmark & \checkmark & \checkmark & \textbf{20.09} & \textbf{20.54} \\
    \bottomrule
    \end{tabular}
    \label{tab:component}
\end{table*}

    For SLT, we compare our MSKA-SLT with state-of-the-art methods on Phoenix-2014T and CSL-Daily as shown in Tab.~\ref{tab:slt}. We achieved BLEU-4 scores of $29.03$ and $25.43$ on the test sets of these two datasets, respectively, marking an improvement of $0.61$ and $0.1$ BLEU-4 scores compared to the keypoint-based methods.
	Furthermore, our approach on the Phoenix-2014T dataset demonstrated a $0.08$ improvement in BLEU-4 score compared to the previous state-of-the-art (SOTA) method.
    
    The results indicate that our MSKA demonstrates significant performance enhancements on SLR and SLT. This highlights the benefit of initially decoupling keypoint sequences for multi-stream attention, followed by aggregating distinct stream feature representations, thereby distinguishing our MSKA from previous SLR and SLT systems.

    \subsection{Ablation Studies}

	\subsubsection{Impact of Keypoint Augment}

    To explore the significance of different data augmentation techniques in SLR endeavors, we methodically implemented each augmentation approach in the training of our models and assessed their efficacy in the context of the Phoenix-2014T SLR challenge. The outcomes are detailed in Table~\ref{tab:augment}. It is discernible that the efficacy of model began to diminish upon integrating of translation and scaling data augmentation methods. We posit that these particular augmentation strategies introduced discrepancies in the data alignment with the validation set, consequently resulting in overfitting. Hence, we made the decision to exclusively employ temporal and rotational augmentations.

    \subsubsection{Impact of Each Component}
    
    We initially demonstrate the impacts of each stream of MSKA-SLR in Table.~\ref{tab:component}. In the absence of the multi-stream architecture, the solitary body stream (where one keypoint attention module manages all keypoints) achieves $25.30$\% and $25.50$\% WER on the Phoenix-2014T. Within Table.~\ref{tab:component}, we present the results separately for the left, face and right streams, as well as the fused outcome. This signifies that the precision of segregated streams is substandard compared to that of the solitary body stream, attributable to the loss of certain information. Nonetheless, owing to the distinct focuses and mutual enhancement among these three streams, their fusion culminates in a WER performance of $23.67$\%, marking a $1.83$\% enhancement over the solitary body stream. To optimize the attributes that our model attends to, we integrate the body stream into the fusion head, resulting in a WER performance of $22.70$\%. Ultimately, by incorporating the self-distillation, our framework achieves the optimal outcome, yielding a WER of $20.54$\%.

    Moreover, in our experiments, we also found that in sign language, the right hand exhibits a more prevailing role compared to the left hand. In our study, the results from only using the left hand and the right hand differ by approximately $22$\%. This discrepancy may be attributed to the fact that in the majority of individuals, the right hand is the dominant hand, while the left hand is the non-dominant hand. Consequently, the right hand is more suitable for performing the detailed and sophisticated gestures essential for sign language. This results in the right hand typically bearing more responsibility and encompassing more information in sign language.

    \begin{table*}[!t]
    \caption{Ablation studies of: (a) methods for augmenting keypoint data; (b) the impact of varying the number of keypoint attention modules; (c) the effects of varying the number of attention heads in an attention module; (d) the weight of the distillation loss; (e) impact of Spatial Global Regularization; (f) effectiveness of the spatial-temporal attention.}
    \centering
    \begin{subtable}[b]{0.4\textwidth}
    \centering
    \resizebox{\linewidth}{!}{
    \begin{tabular}{cccccc}
    \toprule 
    temporal & rotate & translate & scale & Dev & Test \\
    \midrule
    & & & & 24.98 & 25.75 \\
    \midrule
    \checkmark  &            &             &  & 20.25 & 21.58 \\
    \checkmark  & \checkmark &             &  & \textbf{20.09} & \textbf{20.54} \\
    \checkmark  & \checkmark & \checkmark  &  & 20.37 & 21.01 \\
    \checkmark  & \checkmark & \checkmark  & \checkmark &  21.21 & 22.21 \\
    \bottomrule
    \end{tabular} }
    \caption{The combined effects of various data augmentation techniques.}
    \label{tab:augment}
    \end{subtable}
    \hfill
    \begin{subtable}[b]{0.25\textwidth}
    \centering
    \resizebox{\linewidth}{!}{
    \begin{tabular}{ccc}
    \toprule
    Modules  & Dev & Test \\
    \midrule
    6 & 20.79& 21.28 \\ 
    8 & \textbf{20.09} & \textbf{20.54}  \\ 
    10 &  20.87 & 21.53 \\ 
    12 &  22.45 & 22.71 \\
    \bottomrule
    \end{tabular} }
    \caption{The number of attention modules.}
    \label{tab:modules}
    \end{subtable} 
    \hfill
    \begin{subtable}[b]{0.22\textwidth}
    \centering
    \resizebox{\linewidth}{!}{
    \begin{tabular}{ccc}
    \toprule
    Heads  & Dev & Test \\
    \midrule
    2 & 20.97 & 21.42 \\ 
    4 & 20.47 & 21.39 \\ 
    6 & \textbf{20.09} & \textbf{20.54} \\ 
    8 & 20.20 & 21.53 \\
    \bottomrule
    \end{tabular} }
    \caption{The number of attention heads.}
    \label{tab:heads}
    \end{subtable} 
    \newline
    \begin{subtable}[b]{0.25\textwidth}
    \centering
    \resizebox{0.8\linewidth}{!}{
    \begin{tabular}{ccc}
    \toprule
    L & Dev & Test \\
    \midrule
    0.1 & 20.60 & 20.96 \\
    0.5 & 20.12 & 21.08 \\
    1   & \textbf{20.09} & \textbf{20.54} \\
    2   & 20.77 & 21.79 \\
    \bottomrule
    \end{tabular}}
    \caption{The weight of the distillation loss.}
    \label{tab:weight}
    \end{subtable}
    \hfill
    \begin{subtable}[b]{0.25\textwidth}
    \centering
    \resizebox{0.8\linewidth}{!}{
    \begin{tabular}{ccc}
    \toprule
    SGR  & Dev & Test \\
    \midrule
    \XSolidBrush &  21.24 & 21.15 \\
    \checkmark   &  \textbf{20.09} &  \textbf{20.54} \\
    \bottomrule
    \end{tabular}}
    \caption{SGR: Spatial Global Regularization.}
    \label{tab:sgr}
    \end{subtable}
    \hfill
    \begin{subtable}[b]{0.4\textwidth}
    \centering
    \resizebox{\linewidth}{!}{
    \begin{tabular}{cccc}
    \toprule
    Spatial-attn & Temporal-attn & Dev & Test \\
    \midrule
    \checkmark 	&			 &  \textbf{20.09} & \textbf{20.54} \\
    \checkmark  & \checkmark &  25.45          &  25.73 \\
    \bottomrule
    \end{tabular} }
    \caption{Spatial-attnention and temporal-attention.}
    \label{tab:st}
    \end{subtable}
    \newline
\end{table*}

    \subsubsection{Impact of Attention Modules}

    The influence of network depth on model efficacy stands as a pivotal concern within the realm of deep learning. Broadly speaking, increasing the number of network layers may enhance model performance, but it can also lead to overfitting. Consequently, we have deliberated upon the impact of the number of attention modules on model efficacy. We have designated the number of modules as $6$, $8$, $10$ and $12$, as delineated in Table~\ref{tab:modules}. We ascertain that the pinnacle of performance is attained with $8$ modules, yielding the superlative outcome of $20.54$\% WER. Additionally, we have delved into the ramifications of attention heads within the attention module on the network. This facilitates the model to simultaneously assimilate information across diverse representation subspaces. Each head possesses the capability to concentrate on distinct segments of the input sequence, thereby significantly amplifying the model's eloquent capacity and its adeptness in capturing intricate relationships. To scrutinize the significance of the number of heads in keypoint attention, we employ assorted quantities of heads and evaluate their performance in the SLR task, as delineated in Table~\ref{tab:heads}. 

    \subsubsection{Impact of Self-Distillation weight}

    As different streams embody the same meaning, we integrate self-distillation loss at the end of the model to integrate the features learned by each component. It is a hyper-parameter that is designed to balance the effect of CTC loss and the self-distillation loss. We conduct experiments by varying the weight. Table~\ref{tab:weight} shows that our MSKA-SLR attains the best performance when the weight is set to $1.0$.

    \subsubsection{Impact of Spatial Global Regularization}

    SGR operates on the attention maps within the attention module to mitigate overfitting. In our experiments, delineated in Table~\ref{tab:sgr}, we initially attained a performance of $21.15$\% WER on the SLR task without incorporating SGR. Subsequently, through the inclusion of SGR, we achieved the optimal performance of $20.54$\% WER. Moreover, we explored methodologies for managing temporal information in keypoint sequences: 1) reorganizing temporal data via temporal attention post spatial attention, and 2) exclusively employing 2D convolutions devoid of temporal attention. The outcomes are delineated in Table~\ref{tab:st}. It is evident that model achieves WER of $25.73$\% with the inclusion of temporal attention, whereas utilizing only 2D convolutions results in $20.54$\% WER. This could be ascribed to the augmentation in parameter quantity, the comparatively diminutive dataset extent, and the heightened vulnerability of model to overfitting.
    
\section{Conclusion}

In this paper, we concentrate on how to introduce domain knowledge into sign language understanding. To achieve the goal, we present a innovative framework named MSKA-SLR which adopts four streams to keypoint sequences for sign language recognition. A variety of methodologies to make the four streams interact with each other. We further extend MSKA-SLR to a sign language translation model by attaching an MLP and a translation network, resulting in the translation framework named MSKA-SLT. Our MSKA-SLR and MSKA-SLT achieve encouraging improved performance on SLR and SLT tasks across a series of datasets including Phoenix-2014, Phoenix-2014T, and CSL-Daily. We achieved state-of-the-art performance in the Phoenix-2014T sign language translation task. We hope that our approach can serve as a baseline to facilitate future research.

\bmhead{Data Availability}
The Phoenix-2014 and Phoenix-2014T datasets are publicly available at \url{https://www-i6.informatik.rwth-aachen.de/~koller/RWTH-PHOENIX/} and \url{https://www-i6.informatik.rwth-aachen.de/~koller/RWTH-PHOENIX-2014-T/}, respectively. The CSL-Daily datasets will be made available on reasonable request at \url{http://home.ustc.edu.cn/~zhouh156/dataset/csl-daily/}.

\bibliography{sn-bibliography}

\end{document}